\documentclass{article}
\usepackage{spconf,amsmath,graphicx,hyperref}
\usepackage{amsmath}   
\usepackage{amssymb}    
\usepackage{amsfonts}   
\usepackage{cite} 
\usepackage{stfloats}
\usepackage{xcolor}
\usepackage[table]{xcolor}
\usepackage{graphicx}
\usepackage{colortbl}

\title{Beyond Mapping : Domain-Invariant Representations via Spectral Embedding of Optimal Transport Plans}

\name{Abdel Djalil Sad Saoud $^{\dagger}$ \qquad Fred Maurice Ngolè Mboula $^{\dagger}$ \qquad Hanane Slimani $^{\dagger}$}

\address{$^{\dagger}$ Universite Paris-Saclay, CEA, List, F-91120, Palaiseau, France }

\begin{document}
%
\maketitle
\begin{abstract}
\textit{Distributional shifts between training and inference-time data remain a central challenge in machine learning, often leading to poor performance. It motivated the study of principled approaches for domain alignment, such as optimal transport based unsupervised domain adaptation, that relies on approximating Monge map using transport plans, which is sensitive to the transport problem regularization strategy and hyperparameters, and might yield biased domains alignment. In this work, we propose to interpret smoothed transport plans as adjacency matrices of bipartite graphs connecting source to target domain and derive domain-invariant samples' representations through spectral embedding. We evaluate our approach on acoustic adaptation benchmarks for music genre recognition, music-speech discrimination, as well as electrical cable defect detection and classification tasks using time domain reflection in different diagnosis settings, achieving overall strong performances.
}
\end{abstract}

\section{Introduction} \label{sec:intro}
The foundational paradigm of machine learning is based on the assumption that training and test data follow the same underlying probability distribution. This assumption ensures that minimizing empirical risk on the training set leads to a low expected risk on unseen data, allowing robust generalization. In practice, however, this premise is often violated due to distributional shifts, where the joint features-labels distribution changes between training and inference, arising from temporal nonstationarities, heterogeneous acquisition hardware, environmental variations, or sampling bias. Consequently, the decision boundaries learned from the training data may not generalize well to deployment. This limitation motivates the study of \textit{domain adaptation}, which addresses distributional shifts by explicitly reducing the discrepancy between the source distribution \(P_s(X,Y)\) and the target distribution \(P_t(X,Y)\). As shown in \cite{ben2010theory}, the discrepancy between probability measures is a key factor in determining the model's generalizability to unseen domains, which inspires approaches that directly minimize it. Among these, \textit{optimal transport} has proven particularly effective in aligning source and target distributions \cite{courty2017joint}. In fact, it provides a principled and intuitive way to push labeled source samples onto the target domain using the so-called barycentric mapping. However, this mapping relies on transport plans that depend on the optimal transport problem regularization strategy and hyperparameters, and might yield biased domain alignment if not carefully chosen. 
\medskip\par
\textbf{Previous works.} Transfer learning aims at leveraging knowledge from a source task to better learn a similar target task, addressing both scenarios where tasks differ despite similar features distributions, and distributional shifts between domains~\cite{pan2009survey}.
Early research focused on detecting and estimating distribution shifts~\cite{kifer2004detecting}. It was later formalized in~\cite{ben2006analysis, bendavid2010impossibility}, which provided theoretical foundations, both for single-source and multi-source domain adaptation, including generalization bounds and limitations for adapting across domains. Building on this theoretical groundwork, Optimal Transport (OT) has been introduced as a systematic way to align distributions between domains. Initial OT-based methods tackled single-source adaptation by minimizing the Wasserstein distance between source and target distributions~\cite{courty2016optimal, cuturi2013sinkhorn}. Beyond this setting, Joint Class Proportion and Optimal Transport (JCPOT)~\cite{redko2019optimal} considers situations where the target label distribution differs from the sources and adjusts the alignment accordingly. Weighted Joint Distribution Optimal Transport (WJDOT)~\cite{pmlr-v180-turrisi22a} addresses the case where target labels are unavailable, by estimating a weighting vector over the joint distributions of sources. Building further on these ideas, authors in~\cite{montesuma2021wasserstein} construct an intermediate domain by combining source distributions via a Wasserstein barycenter, which is then aligned with the target domain by solving a simple OT problem. The theoretical and computational aspects of Wasserstein barycenters, which underpin these methods, are discussed in~\cite{agueh2011barycenters, cuturi2014fast}. Beyond distribution alignment, some graph-based methods exploit the internal structure of domains: SPA~\cite{xiao2023spa} uses graph spectral alignment to capture intra-domain structures, while LPOT~\cite{el2021label} also builds a bipartite graph from the OT plan but uses it for label propagation. 

\medskip\par
\textbf{Contributions.} Our main contributions are the following: \textit{(1)} we propose a new OT-based domain adaptation framework that leverages the cross-domain connectivity captured by the transport plans to compute domain-invariant and discriminative representation of samples, rather than estimating a mapping from one domain to another in the samples space; \textit{(2)} we propose a new multi-source domain adaptation algorithm in this framework; and \textit{(3)} we evaluate and compare the proposed method with state-of-the-art methods on acoustic adaptation benchmarks, along with a cable defects diagnosis use case, based on Time Domain Reflectometry, to highlight industrial relevance.
\medskip
\par \textbf{Paper Organization.} The rest of the paper is organized as follows: Section~\ref{sec:proposedmethod} provides a detailed description of the proposed approach along with its theoretical background. In Section~\ref{sec:experiments}, we describe the experimental setup, report the results, and provide a comparison with state-of-the-art methods. Finally, Section~\ref{sec:conclusion} concludes this paper.


\section{Proposed Approach} \label{sec:proposedmethod} 

This section introduces our proposed method, \textit{Spectral Embedding of Optimal Transport Plans (SeOT)}, for multi-source domain adaptation setting, after outlining the theoretical foundations of optimal transport, followed by the principles of spectral embedding .

\subsection{Optimal Transport Background} \label{subsec:otbackground}

Optimal Transport is a mathematical theory that formalizes the problem of transporting mass between probability distributions while minimizing a given cost~\cite{villani2009optimal}. In this work, we adopt the discrete formulation of OT for empirical distributions. Let $\mu_s \in \Omega^{n_s}$ and $\mu_t \in \Omega^{n_t}$ be probability measures supported on the source and target samples $X_s = \{x_i^s\}_{i=1}^{n_s} \subset \mathbb{R}^d$ and $X_t = \{x_j^t\}_{j=1}^{n_t} \subset \mathbb{R}^d$, respectively, where $\Omega^n$ denotes the probability simplex in $\mathbb{R}^n$. The \textit{Kantorovich formulation} of the  optimal transport problem consists in solving the following optimization problem
\begin{equation}\label{eq:wassersteindistance}
\min_{\gamma \in \Pi(\mu_s, \mu_t)} \sum_{i,j} C_{ij}\, \gamma_{ij},
\end{equation}

where $C_{ij} = \|x_i^s - x_j^t\|^p$ is the cost of transporting a unit of mass from $x_i^s$ to $x_j^t$, $\Pi(\mu_s, \mu_t) = \{ \gamma \in \mathbb{R}_+^{n_s \times n_t} \mid \gamma \mathbf{1}_{n_t} = \mu_s, \; \gamma^\top \mathbf{1}_{n_s} = \mu_t \}$ is the set of couplings composed of joint probability distributions over the product space, and $\gamma$ is a joint probability matrix whose row and column marginals correspond to $\mu_s$ and $\mu_t$, respectively. The solution $\gamma^*$ defines the optimal transport plan

\begin{equation}\label{eq:otplanemd} 
\gamma^* = \arg\min_{\gamma \in \Pi({\mu}_s, {\mu}_t)} \sum_{i,j} C_{ij}\, \gamma_{ij} = \langle C, \gamma \rangle_F, 
\end{equation} 

with $\langle \cdot, \cdot \rangle_F$ denoting the Frobenius inner product. Solving this linear program scales in $\mathcal{O}(n^3\log n)$, when $n_s = n_t = n$, which makes it impractical for large-scale datasets. Following~\cite{cuturi2013sinkhorn}, an entropic regularization term is added to the OT objective to control the smoothness of the coupling $\gamma$ and improve computational tractability

\begin{equation}\label{eq:sinkwassersteindistance}
\min_{\gamma \in \Pi(\mu_s, \mu_t)} \langle C, \gamma \rangle_F - \varepsilon H(\gamma),
\end{equation} 

where $H(\gamma) = - \sum_{i,j} \gamma_{ij}\log \gamma_{ij}$ is the Shannon entropy of $\gamma$, and $\varepsilon > 0$ is the regularization parameter. This regularization makes the optimization problem strictly convex, ensuring the uniqueness of the solution. The resulting objective can be efficiently minimized using iterative scaling methods~\cite{cuturi2013sinkhorn}, which converges linearly and scales quadratically with the number of samples. Moreover, the entropic bias creates a local connectivity between clusters in the source and target domains by spreading the mass across points within corresponding clusters, a property that our approach explicitly exploits to improve domain alignment.


\subsection{Spectral Graph Embedding}\label{subsec:spectralembedding} 

\begin{figure*}[!t] 
\centering 
\hspace{-0.5cm}\includegraphics[width=0.95\textwidth]{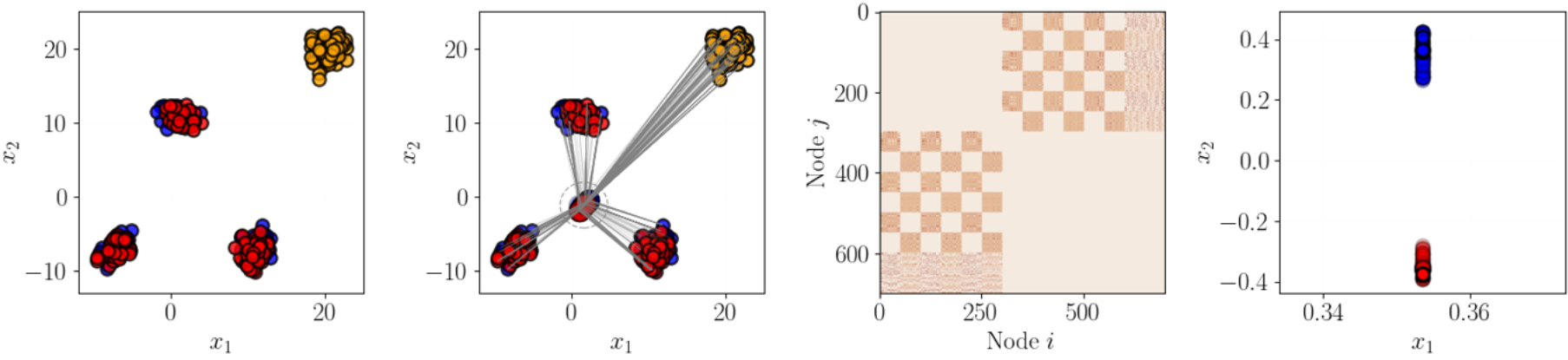} \caption{Illustration of \textbf{SeOT}: (Left) Data from 3 labeled source domains and unlabeled target domain. (center-left) Domains connectivity through Wasserstein barycenter, where the lines connecting different domains refer to the optimal transport plans. (center-right) Constructed adjacency matrix ($\mathcal{A}$) derived from optimal transport plans between domains. (Right) Spectral embeddings of the unified graph representation, demonstrating that class clusters are well-separated in the latent space.} 
\label{fig:placeholder}
\end{figure*} 

Spectral graph embedding aims to represent the vertices of a graph in a Euclidean space while preserving the structural properties of the graph~\cite{chung1997spectral,belkin2003laplacian}. Formally, let $G = (V, E, W)$ be a finite, undirected, weighted graph with $|V| = n$ vertices. Its adjacency matrix $A \in \mathbb{R}^{n \times n}$ is given by

\begin{equation}\label{eq:adjacency} 
A_{ij} = \begin{cases} w_{ij}, & (i,j) \in E,\\ 0, & \text{otherwise}. 
\end{cases} \end{equation}

which encodes connections between vertices, while the \textit{degree matrix}, summarizing each vertex's overall connectivity, takes the form $D = \operatorname{diag}(d_1, \dots, d_n)$, with $d_i = \sum_{j} A_{ij}$. Based on these, the \textit{symmetric normalized Laplacian} is defined as~\cite{chung1997spectral}

\begin{equation}\label{eq:laplacian} 
\mathcal{L}_{\mathrm{sym}} = I - D^{-1/2} A D^{-1/2}. 
\end{equation}
which is positive semi-definite, with eigenvalues $0 = \lambda_1 \le \lambda_2 \le \dots \le \lambda_n$. Importantly, the multiplicity of the eigenvalue $\lambda_1 = 0$ equals the number of connected components $\mathrm{Comp}_1, \dots, \mathrm{Comp}_k$ in the graph, and its corresponding eigenspace is spanned by the vectors $D^{1/2}\mathbf{1}_{\mathrm{Comp}_i}$~\cite{clustering_book}. The spectral embedding is then introduced in terms of $\mathcal{L}_{\mathrm{sym}}$, with the aim of finding orthonormal vectors $f_1, \dots, f_k$ collected in a matrix $F \in \mathbb{R}^{n \times k}$ by solving

\begin{equation}\label{eq:spectral_embedding_eigenvec}
\min_{F \in \mathbb{R}^{n \times k}} \operatorname{Tr}(F^\top \mathcal{L}_{\mathrm{sym}} F) \quad \text{s.t.} \quad F^\top F = I. 
\end{equation}

The solution $F^*$ consists of the eigenvectors of $\mathcal{L}_{\mathrm{sym}}$ associated with the $k$ smallest eigenvalues, where each vertex is represented by a row vector of dimension $k$ in $F^*$,  providing a representation of the graph suitable for nodes clustering. 

\subsection{Spectral Embedding of Optimal Transport Plans}\label{subsec:sewb}
In this work, we interpret the optimal transport plan $\gamma^*$ as an adjacency matrix that captures intrinsic geometric connectivity between domains. To formalize this idea, we first consider the case of two domains and define a bipartite graph weighted by a smoothed OT plan ${\gamma}^*$, estimated by solving Eq.~\ref{eq:sinkwassersteindistance}. We then define the corresponding adjacency matrix as

\[
\hspace{1cm}\mathcal{A}^* = 
\scalebox{1.0}{$
\begin{bmatrix} 
0 & {\gamma}^* \\ 
({\gamma}^*)^\top & 0 
\end{bmatrix} \in \mathbb{R}^{(n_s+n_t) \times (n_s+n_t)}
$}
\]

For multi-source setting \textit{.i.e,} multiple labeled source domains \(\mathcal{D}_s^i = (\mathcal{X}_s^i, \mu_s^i), \, i = 1, \dots, N_s\), and an unlabeled target domain \(\mathcal{D}_t = (\mathcal{X}_t, \mu_t)\),  where $\mathcal{X} \subseteq \mathbb{R}^d$ is the feature space and $\mu$ is a probability measure over it~\cite{courty2017joint,pan2009survey}, we first compute a labeled \textit{Wasserstein barycenter} \(\mathcal{D}_b = (\mathcal{X}_b, \mu_b)\) of the source distributions, by minimizing the weighted sum of Wasserstein distances to all source distributions 
using the algorithm proposed in~\cite{montesuma2021wasserstein}. The multi-source adjacency matrix is then constructed as
\[
\hspace{-0.7cm}\mathcal{A}^* =
\scalebox{0.8}{$
\begin{bmatrix}
0 & \gamma^*_{b \to s_1} & \dots & \gamma^*_{b \to s_{N_s}} & \gamma^*_{b \to t} \\
(\gamma^*_{b \to s_1})^\top & 0 & \dots & 0 & 0 \\
\vdots & \vdots & \ddots & \vdots & \vdots \\
(\gamma^*_{b \to s_{N_s}})^\top & 0 & \dots & 0 & 0 \\
(\gamma^*_{b \to t})^\top & 0 & \dots & 0 & 0
\end{bmatrix}
$}
\]

where $\gamma^*_{b \to s_i}$ and $\gamma^*_{b \to t}$ are entropic transport plans mapping the barycentric distribution to the source distribution $i$ and the target distribution respectively. Thus, \(\mathcal{A}^* \in \mathbb{R}^{K \times K}\), where \(K = n_b + \sum_{i=1}^{N_s} n_s^i + n_t\) is the total number of points across the barycenter, source domains, and target domain, respectively. Zero blocks indicate that all inter-domain connectivity routed through the barycenter. Defining the graph in this manner allows the connected components to reflect cross-domain clusters of samples sharing the same label. Once the adjacency matrix \(\mathcal{A}^*\) is constructed, a spectral embedding of the multi-domain graph is computed as in Section~\ref{subsec:spectralembedding} \textit{i.e,} compute $\mathcal{L}_{\mathrm{sym}}$ according to Eq.~\eqref{eq:laplacian} then Solving Eq.~\eqref{eq:spectral_embedding_eigenvec}. Finally, a classifier is learned using the barycentric nodes for training. 


\section{Experiments}\label{sec:experiments}

\begin{table*}[t] \centering \resizebox{\textwidth}{!}{%
\begin{tabular}{|l|ccccc|c|cccc|c|} \hline Benchmark & \multicolumn{6}{|c|}{MSD} & \multicolumn{5}{c|}{MGR} \\ \hline Domains & Noiseless & Buccaneer2 & Destroyerengine & F16 & Factory2 & Average & Buccaneer2 & Destroyerengine & F16 & Factory2 & Average \\ \hline Source-only $\ddagger$ & 67.99\textsubscript{±8.62} & 82.43\textsubscript{±1.75} & 51.57\textsubscript{±2.56} & 88.89\textsubscript{±2.72} & 50.02\textsubscript{±2.21} & 68.18\textsubscript{±3.47} & 22.90\textsubscript{±0.84} & 38.25\textsubscript{±0.91} & 51.57\textsubscript{±1.11} & 47.80\textsubscript{±0.34} & 40.13\textsubscript{±11.07}  \\ KMM $\ddagger$& 74.64\textsubscript{±6.70} & 87.12\textsubscript{±2.79} & 52.35\textsubscript{±2.94} & 74.86\textsubscript{±5.58} & 50.41\textsubscript{±2.17} & 67.88\textsubscript{±4.04} & 21.75\textsubscript{±0.99} & 39.25\textsubscript{±0.66} & 49.81\textsubscript{±1.69} & 47.37\textsubscript{±0.71} & 39.54\textsubscript{±10.99} \\ TCA $\ddagger$& 50.01\textsubscript{±2.53} & 90.43\textsubscript{±1.40} & 87.14\textsubscript{±4.99} & 95.12\textsubscript{±2.02} & 84.76\textsubscript{±3.30} & 81.49\textsubscript{±2.75} & \underline{58.95\textsubscript{±1.27}} & 60.67\textsubscript{±2.07} & \underline{68.75\textsubscript{±2.11}} & 59.82\textsubscript{±0.50} & 62.04\textsubscript{±3.91} \\ OT-IT $\ddagger$& 89.46\textsubscript{±1.22} & 89.26\textsubscript{±1.56} & 82.84\textsubscript{±2.78} & 84.97\textsubscript{±3.09} & 91.21\textsubscript{±2.04} & 89.76\textsubscript{±2.34} &56.35\textsubscript{±0.84} & \underline{61.92\textsubscript{±1.64}} & 66.72\textsubscript{±1.86} & 61.77\textsubscript{±1.65} & 61.69\textsubscript{±3.67}  \\ OT-Laplace $\ddagger$& 90.44\textsubscript{±1.37} & 87.28\textsubscript{±2.97} & 84.38\textsubscript{±1.76} & 86.14\textsubscript{±2.79} & 90.61\textsubscript{±1.68} & 87.27\textsubscript{±2.11} & 58.02\textsubscript{±1.45} & 60.47\textsubscript{±1.75} & 66.55\textsubscript{±1.60} & 63.87\textsubscript{±1.51} & \underline{62.23\textsubscript{±3.24}}  \\ JCPOT $\ddagger$& 65.66\textsubscript{±5.71} & 92.55\textsubscript{±2.11} & 87.89\textsubscript{±1.39} & 88.67\textsubscript{±1.67} & 82.41\textsubscript{±2.22} & 83.44\textsubscript{±2.62} &35.87\textsubscript{±0.41} & 48.47\textsubscript{±2.97} & 51.92\textsubscript{±3.25} & 51.95\textsubscript{±1.75} & 47.05\textsubscript{±6.60} \\ JCPOT-LP $\ddagger$& 12.89\textsubscript{±1.67} & 89.06\textsubscript{±1.38} & 84.97\textsubscript{±3.23} & 90.24\textsubscript{±1.71} & 86.13\textsubscript{±1.88} & 72.66\textsubscript{±1.97} &36.40\textsubscript{±0.39} & 52.92\textsubscript{±1.32} & 56.30\textsubscript{±0.37} & 51.52\textsubscript{±2.28} & 49.28\textsubscript{±7.62} \\ WBT $\ddagger$& 52.74\textsubscript{±3.82} & 56.88\textsubscript{±9.54} & 56.63\textsubscript{±6.88} & 56.63\textsubscript{±6.56} & 59.38\textsubscript{±2.61} & 58.56\textsubscript{±4.80}  &21.37\textsubscript{±2.25} & 24.30\textsubscript{±2.71} & 25.30\textsubscript{±6.02} & 22.70\textsubscript{±2.25} & 23.41\textsubscript{±1.50} \\ WBT$_{\text{reg}}$ $\ddagger$& \underline{94.34\textsubscript{±2.55}} & \underline{96.27\textsubscript{±1.60}} & \underline{92.98\textsubscript{±1.38}} & \underline{94.92\textsubscript{±0.68}} & \underline{96.87\textsubscript{±0.94}} & \underline{95.08\textsubscript{±1.43}} & \textbf{70.60\textsubscript{±1.27}} & \textbf{83.05\textsubscript{±0.97}} & \textbf{84.40\textsubscript{±1.71}} & \textbf{90.17\textsubscript{±0.46}} & \textbf{82.05\textsubscript{±7.13}} \\ Target-only $\ddagger$& 96.88\textsubscript{±2.97} & 90.51\textsubscript{±3.98} & 93.07\textsubscript{±3.81} & 89.23\textsubscript{±4.25} & 92.30\textsubscript{±3.62} & 92.40\textsubscript{±3.73}  &67.43\textsubscript{±1.43} & 67.96\textsubscript{±2.91} & 66.86\textsubscript{±2.00} & 68.37\textsubscript{±1.87} & 67.41\textsubscript{±0.56} \\  \hline \rowcolor{gray!15} \textbf{SeOT} & \textbf{99.22\textsubscript{±0.00}} & \textbf{96.61}\textsubscript{±0.97} & \textbf{97.40\textsubscript{±0.37}} & \textbf{95.31\textsubscript{±0.00}} & \textbf{98.70\textsubscript{±0.37}} & \textbf{97.45\textsubscript{±0.34}}  & 45.53\textsubscript{±0.12} & 61.63\textsubscript{±0.31} & 58.17\textsubscript{±0.12} & \underline{70.77}\textsubscript{±0.19} & 59.03\textsubscript{±0.19} \\ \hline 

\end{tabular}}
\caption{Classification results on the MSD and MGR benchmarks. $\ddagger$ denotes results reported from~\cite{montesuma2021wasserstein}, except for the Noiseless domain and the overall average across all domains for MSD. Each column shows results with that domain as target, with values reported as mean accuracy ± standard deviation, averaged over three independent runs.} 
\label{tab:results_benchmarks} 
\end{table*}

\subsection{Experimental setup}\label{subsec:datsets} 
\textbf{Datasets.} To evaluate our proposed methods alongside existing approaches, we consider three datasets: \textit{(1)} The Music-Speech Discrimination dataset \textit{(MSD)}~\cite{tzanetakis2001musical} contains 64 music and 64 speech excerpts, used for binary classification across five noise domains: \textit{Noiseless}, \textit{Buccaneer2}, \textit{Destroyerengine}, \textit{F16}, and \textit{Factory2}. \textit{(2)} The Music Genre Recognition dataset \textit{(MGR)}~\cite{tzanetakis2001musical} comprises 1,000 recordings across ten genres, used for multi-class classification under varying background noise conditions, and \textit{(3)} \textit{CS-RT} Cables fault diagnosis using time domain reflectometry and compressive sensing~\cite{slimani2024detection}. 
Here, Four domains are considered consisting of three different compression factors of reflectometric signals (16, 4 and 2) with the same simulated coaxial cable, and reflectometric signals from a cable with different electrical features, denoted "Phys". Each domain includes 200 signals, each represented by 512 features and belonging to one of the following classes: no fault, soft fault, short circuit and open circuit.
\medskip\par
\textbf{Classifiers.} For MSD and CS-RT, we used a four-layers MLP (d $\rightarrow$ 512 $\rightarrow$ 512 $\rightarrow$ 512 $\rightarrow$ 512 $\rightarrow$ $N_c$), with ReLU activations, trained with batch size 128 using Adam optimizer (lr = $10^{-3}$, and weight decay = $10^{-5}$). 
For MGR, a Random Forest with 1000 trees and maximum depth 13 was employed.

\subsection{Results}\label{subsec:results}

\textbf{Main results.} The results on the MSD and MGR benchmarks are reported in Table~\ref{tab:results_benchmarks}. On MSD, SeOT improves over the source-only
baseline by nearly 29\% on average and outperforms all other methods across all domains, showing clear and consistent gains. Remarkably, the performance of SeOT is even higher than the target-only case, where one assumes that labeled
target data is used to train a classifier. As reported in \cite{montesuma2021wasserstein}, MGR is more challenging since it involves more classes, which yields more class mismatches in the transport plan between the barycentric distribution and the target, when no class information is leveraged for WBT and SeOT. Although WBT$_{reg}$ achieves the best overall accuracy, SeOT still improves over the source-only scenario by more than 18\% on average, contrary to WBT, highlighting the benefit of our spectral approach. The results on the CS-RT benchmark are presented in Table~\ref{tab:results_reflecto}. Here, SeOT improves the average performance over the source-only baseline by almost 25\%, while competing methods failed to achieve any notable gain on average for this dataset.


\begin{table}[t] 
\centering 
\resizebox{\columnwidth}{!}{%
\begin{tabular}{|l|cccc|c|} \hline Benchmark & \multicolumn{5}{c|}{CS-RT} \\ \hline Algorithm & CF$_{16}$ &  CF$_4$ & CF$_{2}$ & Phys. & Average \\ 
\hline Source-only & 23.00\textsubscript{±0.00} & 28.00\textsubscript{±0.00}  & 51.00\textsubscript{±0.00} & 47.00\textsubscript{±0.00} & 37.25\textsubscript{±0.00} \\ 
KMM & 23.00\textsubscript{±0.00} & 25.00\textsubscript{±0.00}  & 45.00\textsubscript{±0.00} & 44.50\textsubscript{±0.00} & 34.38\textsubscript{±0.00} \\ OT-Laplace & 18.00\textsubscript{±0.00}  & 25.50\textsubscript{±0.00} & 46.50\textsubscript{±0.00} & \underline{54.50}\textsubscript{±0.00} & 36.12\textsubscript{±0.00} \\
JCPOT & \underline{29.50}\textsubscript{±0.00}  & \underline{30.00}\textsubscript{±0.00} & 30.50\textsubscript{±0.00} & 41.00\textsubscript{±0.00} & 32.75\textsubscript{±0.00} \\ JCPOT-LP & 24.50\textsubscript{±0.00} & 25.00\textsubscript{±0.00}  & 26.00\textsubscript{±0.00} & 4.50\textsubscript{±0.00} & 20.00\textsubscript{±0.00} \\ 
WBT & 20.00\textsubscript{±0.71}  & 19.67\textsubscript{±1.55} &  31.00\textsubscript{±2.86} & 30.00\textsubscript{±6.48} & 23.75\textsubscript{±2.32} \\ 
WBT$_{reg}$ & 28.83\textsubscript{±0.24} & 16.17\textsubscript{±0.24}  & \underline{52.17}\textsubscript{±3.88} & 38.50\textsubscript{±2.55} & 33.92\textsubscript{±1.73} \\ \hline \rowcolor{gray!15} \textbf{SeOT} & \textbf{49.90\textsubscript{±3.72}}  & \textbf{62.88\textsubscript{±1.17}} & \textbf{65.90\textsubscript{±0.64}} & \textbf{69.59\textsubscript{±5.70}} & \textbf{62.07\textsubscript{±2.25}} \\
\hline 
\end{tabular}} 
\caption{Classification accuracy results for Compressed Sensing Reflectometry Dataset. Each column shows results of that domain as target, averaged across three independent runs.} 
\label{tab:results_reflecto} 
\end{table} 
\par
\begin{figure}
\centering
\vspace{-0.1cm}
\includegraphics[width=0.7\linewidth]{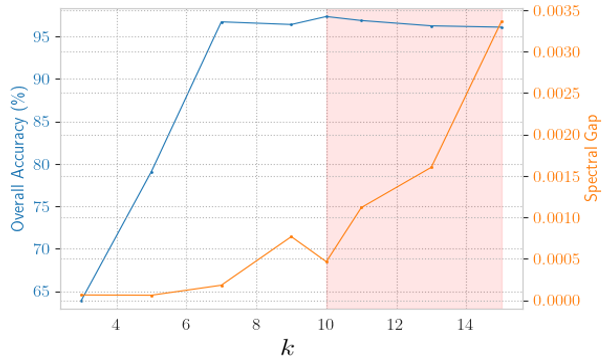}
\caption{The effect of embedding size $k$ on the spectral gap and the average classification accuracy for MSD dataset.}
\label{fig:hyperparameters}
\end{figure}

\textbf{Embedding dimensions.}  We propose an intuitive way to select the embedding dimension $k$ by maximizing the spectral gap between the $N_c$-th and $(N_c+1)$-th smallest eigenvalues. Figure~\ref{fig:hyperparameters} shows the average classification accuracy and spectral gap as a function of $k$, for a fixed $\varepsilon = 10^{-4}$. As $k$ increases, both spectral gap and classification accuracy tend to increase, where the maximum spectral gap is situated within the plateau region of accuracy, which is consistent with the principle that a large gap at that position indicates $N_c$ well-separated connected components in the cross-domain graph, and therefore, more discriminative embeddings. This also provides a principled way to select the entropic regularization parameter $\varepsilon$, which is otherwise set based on a tricky trade-off between convergence speed and entropic bias.
\par
\textbf{Computational complexity.} Our approach involves two steps: (1) solving the entropic OT problem, which scales as $\mathcal{O}(n^2)$, and (2) the eigen-decomposition of the Laplacian, which usually scales as $\mathcal{O}(n^3)$. However, due to the block-sparse structure of $\mathcal{A}$ and the fact that only $k$ eigenvectors are required, we use an Arnoldi iterative solver, for which the cost can be decomposed into: the matrix-vector multiplications with the Laplacian, scaling as $\mathcal{O}(k \mathrm{nnz}(\mathcal{A}))$ \textit{i.e.,} roughly $\mathcal{O}(k k_d n^2)$ for $k_d$ domains of $n$ atoms each, and the orthogonalization of the $k$ vectors, which contributes an additional $\mathcal{O}(k^2 k_d n)$.

\section{Conclusion}\label{sec:conclusion}

In this paper, we proposed \textit{SeOT}, a multi-source domain adaptation method for addressing distributional shifts between source and target domains. Unlike existing approaches that use the optimal transport plan to push labeled source samples onto the target distribution, SeOT uses transport plans to define a graph connecting different domains and performs spectral embedding to derive domain invariant and discriminative samples representation. We evaluated SeOT on the MSD and MGR benchmarks, covering speech and music classification tasks, as well as on the CS-RT dataset, which reflects realistic sensing conditions and practical application scenarios of signal processing. Overall, SeOT achieved competitive performances and notably outperformed competing approaches on the CS-RT dataset.
Future works will include out-of-sample data embedding.

\section*{Acknowledgements}
The authors acknowledge Yosra Gargouri and Nicolas Ravot for their valuable insights and perspectives shared through discussions.

\bibliographystyle{IEEEbib}
\bibliography{strings,refs}

\end{document}